\documentclass{article}

\usepackage{microtype}
\usepackage{graphicx}
\usepackage{subfigure}
\usepackage{booktabs} 

\usepackage{hyperref}

\usepackage[accepted]{icml2024} 

\usepackage{icml2024}

\newcommand{\documentname}{\textsl{Position Paper}}
\renewcommand{\paragraph}[1]{\noindent\par\textbf{#1}}
\newcommand{\sectionname}{Section}
\newcommand{\secref}[1]{\sectionname~\ref{#1}}
\newcommand{\figref}[1]{\figurename~\ref{#1}}
\newenvironment{hoggnumerate}
  {\begin{list}
    {\arabic{enumii}.}
    {\usecounter{enumii}
     \setlength{\itemsep}{0in}
     \setlength{\parsep}{0in}
     \setlength{\parskip}{0in}
    }
  }
{\end{list}}

\icmltitlerunning{Hogg \& Villar: Is machine learning good or bad for the natural sciences?}
\begin{document}

\twocolumn[
\icmltitle{\textsl{Position:} Is machine learning good or bad for the natural sciences?}

\begin{icmlauthorlist}
\icmlauthor{David W. Hogg}{ccpp,mpia,flatiron}
\icmlauthor{Soledad Villar}{flatiron,ams,minds}
\end{icmlauthorlist}
\icmlaffiliation{ccpp}{Center for Cosmology and Particle Physics, Department of Physics, New York University, USA}
\icmlaffiliation{mpia}{Max-Planck-Institut f{\"u}r Astronomie, Heidelberg, Germany}
\icmlaffiliation{flatiron}{Flatiron Institute, New~York, USA}
\icmlaffiliation{ams}{Department of Applied Mathematics and Statistics, Johns Hopkins University, USA}
\icmlaffiliation{minds}{Mathematical Institute for Data Science, Johns Hopkins University, USA}
\icmlcorrespondingauthor{David W. Hogg}{david.hogg@nyu.edu}

\icmlkeywords{Machine Learning, physics, science, philosophy, bias}

\vskip 0.3in

\begin{abstract}
  Machine learning (ML) methods are having a huge impact across all of the sciences.
  However, ML has a strong ontology---in which only the data exist---and a strong epistemology---in which a model is considered good if it performs well on held-out training data.
  These philosophies are in strong conflict with both standard practices and key philosophies in the natural sciences.
  Here we identify some locations for ML in the natural sciences at which the ontology and epistemology are valuable.
  For example, when an expressive machine learning model is used in a causal inference to represent the effects of confounders, such as foregrounds, backgrounds, or instrument calibration parameters, the model capacity and loose philosophy of ML can make the results more trustworthy.
  We also show that there are contexts in which the introduction of ML introduces strong, unwanted statistical biases.
  For one, when ML models are used to emulate physical (or first-principles) simulations, they amplify confirmation biases.
  For another, when expressive regressions are used to label datasets, those labels cannot be used in downstream joint or ensemble analyses without taking on uncontrolled biases.
  The question in the title is being asked of all of the natural sciences; that is, we are calling on the scientific communities to take a step back and consider the role and value of ML in their fields; the (partial) answers we give here come from the particular perspective of physics.
\end{abstract}
] 
\printAffiliationsAndNotice{}

\section{Introduction}\label{sec:intro}
It is an understatement to say that machine learning (ML) is having a big impact across the sciences.
A significant fraction of all scientific papers in the natural sciences now employ ML in part (or all) of their analyses.
(We will define ML below in \secref{sec:philosophy}).
However, when we ask what scientific breakthroughs have been enabled by this influx of new tools and methods, there isn't a long list.
The success of the \textsl{AlphaFold} projects in protein structure \cite{alphafold} are often raised.
But these are successes in a very specific challenge-problem setting in which \emph{performance} is valued over \emph{understanding}.
In the natural sciences we almost exclusively care about understanding, in the long run.

The natural sciences are concerned with understanding the world, and naturally occurring mechanisms in play in that world.
We make progress by discovering new kinds of objects and phenomena, and explaining (and, even better, predicting) qualitatively new kinds of objects and phenomena.
Our most successful investigations are judged in terms of the questions they answer, or the new questions they raise, or both.
The question here is: How will ML contribute to this mission?

In contrast to natural science, ML research and ML methods are concerned with making accurate predictions for, or descriptions of, \emph{data}.
A ML method is considered successful if it performs well on held-out training data, even if the latent structure of the model is generic and the internals are impossible to interpret.
In ML, the considerations are almost all at the level of the data, and we are happy to use models in which we have little or no understanding of the meanings or values of the latent parameters or weights.

In natural science, on the other hand, the most important contributions and results are all at the level of the \emph{latents}:
We use data to learn about the latent structure of the world or of the system we are studying.
In astrophysics, this could be the interior structure of the Sun, or the processes that form planets around other stars, or the map of the dark matter surrounding the Milky Way.
The things we care about are almost never directly observable; they are parameters (or hyper-parameters) of a physical (or chemical or biological) model that predicts the observables.
Often the thing we care about is the model itself.

For a concrete example, when the expansion of the Universe was discovered \cite{expansion, expansion2}, the discovery was important, but not because it permitted us to predict the values of the redshifts of new galaxies (though it did indeed permit that).
The discovery was important because it told us previously unknown things about the age and evolution of the Universe, and it confirmed a prediction of general relativity, which is a theory of the latent structure of space and time.
The discovery would not have been seen as important if Hubble and Humason had instead announced that they had trained a deep multilayer perceptron that could predict the Doppler shifts of held-out extragalactic nebulae.

For another example, consider the discovery that the paths of the planets are ellipses, with the Sun at one focus \cite{kepler}.
This discovery led to extremely precise predictions for data.
It was critical to this discovery that the data be well explained by the theory.
But that was not the primary consideration that made the nascent scientific community prefer the Keplerian model.
After all, the Ptolemaic model preceding Kepler made equally accurate predictions of held-out data.
Kepler’s model was preferred because it fit in with other ideas being developed at the same time, most notably heliocentrism \cite{copernicus} and Newtonian gravity \cite{newton}.

We are both ML \emph{skeptics} and ML \emph{practitioners}.
We are writing this for an audience that is either developing an ML method for science applications or else bringing ML into a scientific domain.
\textbf{
The main points of this \documentname{} are \textsl{(1)}~that ML has places at which it is very valuable in the contemporary practice of science, \textsl{(2)}~that ML also has places at which it will create problems for science, and \textsl{(3)}~currently it is not obvious what is the epistemic role of ML within the broader goals of the natural sciences.
This paper is a call to natural-science communities to think about it, and answer the question in the title for themselves.}
From the perspective of the physical sciences, our answer is ``Both.''

\paragraph{Our contributions:}
\vspace{-1ex}\begin{itemize}
  \item We deliver a description of the fundamental \emph{ontology} and \emph{epistemology} of ML, and contrast these with the ontologies and epistemologies of the natural sciences.
  \item We elucidate two important and strong statistical biases that are being introduced to the natural sciences by some uses of ML. One is a confirmation bias that arises when simulations are replaced or augmented by emulators. Another is a more standard estimator bias that is (possibly enormously) amplified when elements of datasets produced by ML regressions are used in combination or joint or ensemble analyses. Neither of these two biases can be easily corrected.
  \item We show that there are many safe places for the use of ML methods in current natural-science practices, and places where (in the contemporary context) the use of ML methods is effectively \emph{required}. Many of these places are in the operational parts of scientific projects.
  \item We argue that in causal contexts, in which the ML method is used to model foregrounds, backgrounds, calibration parameters, or other confounders, using the most expressive ML method can deliver \emph{the most conservative approach} to the scientific problem.
\end{itemize}

\section{The ontology and epistemology of machine learning}\label{sec:philosophy}

\paragraph{What is machine learning?}
For the purposes of this \documentname, we will employ an expansive or inclusive definition of ML.
For us, a method is ML if its \emph{capability increases substantially as it sees more data} \cite{ml_definition}.
In some sense this definition could be seen as true of any measurement process, since measurements improve as the data improve. 
If we are going to be more specific, we'd like the model precision or capability to improve faster (in some sense) than (something like) the square root of the increment in data.

This definition is broad, perhaps controversially broad:
In addition to methods like convolutional neural networks \cite{cnn}, multi-layer perceptrons, and transformers \cite{transformer}, it includes large linear regressions, Gaussian processes \cite{gp}, support-vector machines \cite{svm}, principal components analysis \cite{pca}, kernel density estimates \cite{kde}, and even some kinds of multilayer or hierarchical models \cite{multilevel}; indeed almost any models in any current ML textbooks (eg, \citealt{ml_book1, ml_book2}).
Feel free to personally take any more restrictive definition.
Our comments will apply to everything in the larger class.

As broad as this definition is, it does not include standard model fitting or parameter estimation with a well-specified (first-principles, say) model.
These tasks are usually performed in the context of a rigid model, with far fewer parameters than data points, in which the model does not qualitatively change as more data arrive, and the quantitative improvements to the parameter estimates only scale as something like the square root of the size of the data set.
Standard model fitting and parameter estimation in the enormously under-parameterized regime are not ML in our sense.

\paragraph{What is natural science?}
For the purposes of this \documentname, we will employ a restrictive definition of natural science.
For us, the natural sciences are the sciences that study the natural world, with the primary aim of \emph{understanding}.
Examples include physics, chemistry, biology, Earth science, ecology, and so on.
For our purposes, the natural sciences are about understanding observed phenomena, unifying knowledge, making predictions for new experiments and observations.
Natural-science questions include things like:
How did our Solar System form?
Or how do cells differentiate as the embryo grows?
Or what causes variations in the jet stream?

For the purposes of this \documentname{} we will exclude more engineering-oriented questions such as: What protein sequences might be important for the treatment of diabetes? Or where in Canada might there be new uranium deposits? Or what kinds of catalytic surfaces might help turn biomass into biofuel?
This is a bit unfair!
We are excluding these kinds of questions in part because it is obvious that ML methods \emph{can} serve important purposes in such problems.

In addition, we recognize that there are quite a few different attitudes that a scientist can take towards the concept of ``understanding''.
Does a program that can predict and control a robot arm accurately ``understand'' the arm?
Or does a system that can predict protein foldings ``understand'' protein folding?
We are using a restricted definition of ``understanding'' that excludes these operational successes; mostly in the natural sciences we are looking for semantic rather than operational understandings of systems.
But of course ML often \emph{can} provide extremely operationally capable systems.

\paragraph{Ontologies:}
The ontology of a domain is (for our purposes) the set of things that exist within that domain.
What is the ontology of ML?

Unsupervised and self-supervised ML methods deliver representations, descriptions, or compressions of data.
Supervised ML methods find relationships between data (features) and data (labels).
In both cases, the methods are judged on their capability to accurately describe the data.
They are (often) not judged on the details of their latent structure.
In a very important sense, the ML ontology is that \emph{only the data exist}.

In support of this point of view, here are some comments:
Deep-learning models have enormously large internal degeneracies (combinatorial degeneracies even).
These degeneracies are not seen as a problem, since all they do is make the latent weights less interpretable without hurting performance \cite{belkin2019reconciling, bartlett2020benign}.
Contemporary optimization schemes are stochastic \cite{stochastic, adam} and most models are non-convex.
The fact that it is impossible to find the global optimum---and the fact that in most contemporary models that isn't even the goal---shows that the latent parameters are not important.
The regularization of trained models with early stopping \cite{early_stop} and dropouts \cite{dropout} demonstrate that there isn't even a goal of getting precisely to \emph{any optimum}.
Any practitioner using a 23-layer network can trivially (and without analysis) switch to a 42-layer, despite the fact that (from a functional point of view), this might substantially change the model capacity or expressivity.

In ML, models are deemed stably optimized and useful not if the latent space---the joint values of all the weights and biases---is stable, but rather if the predictions for held-out data are stable.
This is used as an indicator that the \emph{learned function} (at least in the training, validation, and test set) is stable, even though there can be large subspaces of very different, unidentifiable latent parameters.
Since the learned function is a model for the data, or something that predicts the data, it is the data that exist in ML contexts, not the latent parameters.

In ML there is sometimes additionally a notion of ``ground truth''; does this concept extend the ontology of ML?
Some ML methods assume that the training (and validation) data constitute ground truth (the labels are correct and have knowledge of the world, say).
In this case, ``ground truth'' is just another name for the training data.
Other methods assume that the labels might be noisy or adversarial, and that the method discovers true or more true labels.
This does perhaps extend the ontology, but only slightly, since the ground truth is in the same space as the data (and nearby to it, presumably).
Another notion that sometimes arises in ML is the ``data-generating process'', especially in contexts of emulation.
We don't see this as different from the data themselves, but technically this does extend the ontology marginally.

In contrast to ML, the ontologies of the natural sciences contain far more things than just the data.
In physics, for example, 
not only do the data exist, but so do forces, energies, momenta, charges, spacetime, wave functions, virtual particles, and much more.
These entities are judged to exist in part because they are involved in the latent structure of the successful theories; almost none of them are direct observables.
The most important discoveries in the natural sciences are discoveries of latent structure: Natural selection as an explanation for the differentiated properties of species \cite{natural_selection}, for example, or the elemental composition of the Sun (see \citealt{sun_composition}), or the quarks and gluons that make up the proton \cite{proton_substructure}.

The mismatch of ontology (and epistemology, below) between ML and natural sciences is not unique to ML;
many tools used in the natural sciences have limited ontologies.
Perhaps the most extreme case is that of mathematical tools, which are based (in some sense) only on a limited set of axioms; mathematics has a very limited ontology.
ML is useful in science in a way that is analogous to other methodological tools; the point made here is that the particular and specific use of each tool ought to be matched to its ontology and epistemology.

\paragraph{Epistemologies:}
The epistemology of a domain is (for our purposes) the method or standard by which something is judged to be true or correct or known.
What is the epistemology of ML?

A trained ML model is deemed successful or correct if \emph{it performs well on held-out training data}.
This epistemological position is related strongly to the ontological position:
If only the data exist, then the success of a model is judged only in terms of the data it describes.

In support of this position we could point to the literature on adversarial attacks (eg, \citealt{adversarial1}).
This literature shows us, in dramatic ways, that ML methods are not doing what we (na\"ively) believe that humans or scientists or scientific theories do in comparable circumstances.
And yet, these attacks do not suggest (to most practitioners of ML) that the methods are wrong or require revision.
Even the responses to attacks have responded in ways that involve augmentation of data \cite{adversarial_training}, such that the vulnerability to attack is lessened without compromising the fundamental epistemological point that performance on data is the primary standard of truth.

A critic of our ML epistemology claim could point to the large literature on out-of-sample generalization, transfer learning \cite{transfer}, or the more grand contemporary idea of foundation models \cite{foundation}.
But even in these contexts, models are deemed successful when they explain new or held-out data; the latent structure of the models is not a primary consideration when the validity of the model is in question.

Another critic might point to the literatures on interpretable ML \cite{interpretable} and explainable ML \cite{explainable}.
In these ML subfields, in some cases, attempts are made to understand the internal structure or the behavior of the ML model.
There is not consensus here, but if there were, it might create subfields in which the epistemology of ML gets more complex; maybe in the future some ML models will have to perform well on data \emph{and} be explainable in some important sense.
That would have implications for the epistemology, and probably also the ontology of ML.

In the natural sciences, in contrast, the epistemologies are much more restrictive and much more demanding than they are in ML.
The detailed epistemological framework depends on the natural-science field, and even on the practitioner within the field.
However, in all cases, it is more demanding:
A theory or explanation has to do much more than just explain the data in order to be widely accepted as true.
In physics for example, a model---which, as we note, is almost always a model of latent structure---is judged to be good or strongly confirmed not only if it explains observed data.
It ought to explain data in multiple domains, and it must connect in natural ways to other theories or principles (such as conservation laws and invariances) that are strongly confirmed themselves.
(It is worthy of note here that a successful theory in the natural sciences is usually a combination of a more fundamental theory of the latents, and a less fundamental or auxiliary observation model, which explains how the latents appear in or affect the observable data.)
General relativity \cite{gr} was widely accepted by the community (and very quickly; \citealt{peebles}) not primarily because it explained anomalous data (although it did explain some); it was adopted because, in addition to explaining (a tiny bit of new) data, it also had good structure, it resolved conceptual paradoxes in the pre-existing theory of gravity, and it was consistent with emerging ideas of field theory and geometry.

A recent position piece \citep{donoho2024data} points out that an important aspect of ML's epistemology might be its (aspirational) \emph{reproducibility}.
Namely, many contributions to the field permit anyone to reproduce the results and build on top of them; the idea is that
this aspect of ML is part of the reason that the field moves so fast.
In contrast, it is often argued that the sciences face a reproducibility crisis \cite{baker2016reproducibility};
the (only) positive spin to put on that is that reproducibility in the natural sciences is not nearly as straightforward as it is in ML.
This reproducibility discussion is slightly out-of-scope here; we refer the reader to Donoho's piece and the follow-up commentaries (for example, \citealt{Freire2024Singularity}, \citealt{Milanfar2024Data}).

\section{Why do we need machine learning in the natural sciences?}\label{sec:why}
In this \sectionname{} we give some idea of how and why ML has had such a big impact in the natural sciences, and how and why many scientific projects require ML technologies. 
In most cases, the ML is required on the engineering or execution side of projects that are large-scale in some sense.
The descriptions here are non-exhaustive, they have a strong physics bias to them, and they do not address the many technical nuisances that make this discussion approximate and less specific than we would like.
Our hope is that this \documentname{} is seen as a conversation starter and not as providing a complete answer to any of the posed questions.  

\paragraph{Label transfer:}
Sometimes a project has informative data on a very large number of objects, but precise labels for only a few, maybe obtained through very careful analysis or external data.
In this case, if it is extremely expensive to label more objects, a regression can be trained on the few labeled data points and then the trained regression used to label all the rest.
Below in \secref{sec:bad} we argue that this is not an advisable use of ML in the natural sciences if the expectation is that the regression-generated labels are going to be used certain straightforward ways in downstream scientific projects.

\paragraph{Classification:}
This is the same use case as label transfer, but in the specific case in which the labels are drawn from a finite (small, actually) set of discrete values.
It is subject to the same kinds of issues as label transfer.
  
\paragraph{Speeding up decisions:}
Many scientific projects must make decisions very fast in real time.
The most extreme examples of this are in particle physics, where detectors (such as the \textsl{Atlas} Experiment \cite{atlas} at the \textsl{Large Hadron Collider}) must decide whether to trigger a data-saving event in a tiny fraction of second.
It is often the case that trained ML classifiers can reproduce the selection boundaries as well---or nearly as well---as first-principles models, but with far less computation.
It is critical, if ML methods are used for real-time operations, that the methods and all their latent parameters (weights) be preserved for analysis, operations simulations, and conterfactual exercises, often performed long after the data are taken.

\paragraph{Speeding up simulations:}
In much of the natural sciences, the theory in play is a computation or simulation or digital twin.
These simulations tend to be very computationally expensive, since they often span large ranges of spatial or time scales.
ML regressions can be trained to emulate the simulations, or patch up low-resolution simulations to higher resolutions.
Below in \secref{sec:bad} we argue that the introduction of ML emulators can introduce an unwanted, strong confirmation bias.

\paragraph{Modeling nuisances:}
In most natural-science domains, the quantities of interest are not directly observable, but rather the model or prediction for the data is a combination of the theory of relevance plus auxilliary theories of foregrounds, backgrounds, instrumentation, and noise sources.
When the goal is for these nuisances to be effectively modeled but not necessarily understood in detail, ML approaches can be effective and, as we argue in \secref{sec:good}, even conservative.

\paragraph{Propose objects, materials, or interactions for follow-up:}
In most scientific settings, discoveries are valid if they can be experimentally verified, regardless of the process used to produce the scientific hypothesis.
That is, hypotheses can be generated by creative processes that don't need to be understood, provided that the hypotheses so generated can be tested by experimental methods that \emph{are} well understood.
ML-based generative models can be used to conjecture the existence of objects or materials with certain properties, that can be later verified in lab assays.
One example of this is ML-based drug discovery \cite{kang2018conditional}. 

\paragraph{Outlier detection:}
In many of the sciences, discovery of new or previously unknown phenomena or objects can be of great importance.
A recent example in astronomy is the discovery of fast radio bursts \cite{frbs}, which started as outliers in the data taken for imaging projects and have turned out to be interesting astrophysical objects.
Since expressive unsupervised ML methods can (with some caveats) describe accurately a complex distribution of data, they can also be used to identify (some) rare data points that are unlike any other elements of the training data.

\paragraph{Information questions:}
If a ML method is correctly trained to predict labels $y$ from data $x$, and it succeeds (better than random, say), then it shows that data $x$ contain information within it about labels $y$.
Regressions thus can be used to make one-sided measurements of quantities of information-theoretic interest.

\paragraph{Making discoveries?}
There is a new hope in the sciences that sufficiently trained or constrained models might lead to new insights about scientific theories or might effectively make fundamental discoveries.
There are approaches along the direction of symbolic regression \cite{symbolic1, symbolic2}, and there are approaches along the lines of foundation models \cite{foundation1}.
So far this hope has borne no important new discoveries, but discoveries are not inconceivable.
Indeed, there is a very real sense in which the motivations underlying foundation models---finding generally useful latent representations for qualitatively diverse data---are similar to the motivations underlying theoretical physics.

\medskip
This list isn't exhaustive!
For example, it doesn't mention the use of ML to find summary statistics for the comparison of simulations and data in (say) simulation-based inferences \cite{sbi, simbig}.
It also doesn't mention using ML to emulate likelihoods or likelihood ratios \cite{biwei, likelihood_ratio}.
The uses of ML in the natural sciences are legion \cite{wang2023scientific,zhang2023artificial}.

The short summary of all this is that we need ML in the natural sciences; we can't live without it.
The question here is---given the restrictive ontology and loose epistemology---how we will use it safely?

\section{When is machine learning bad for natural science?}\label{sec:bad}
In this \sectionname{} we elucidate two statistical biases that can be introduced into a natural-sciences project when ML methods are introduced.
Neither of these biases can be easily corrected or removed, to our knowledge.

\paragraph{Amplifications of training-set biases:}
The outcomes of ML regressions are label estimates that are conditioned on the input features \emph{and also} on the totality of the training set used to train the regression.
This is good; the individual-data-point label estimates from the regression are the lowest ``risk'' (in the statistical sense) when they use non-zero bias in the bias--variance trade-off.
However, the biases that are weak or manageable on an individual data-point basis become strong or unacceptable biases when outputs of regressions are used jointly or in combination to measure a population or sub-population or ensemble property.

This problem is demonstrated with a toy example in \appendixname~\ref{app:toy}.
It is worst when the regression is used to label a large data set or catalog or sample of data,
and when elements of that data set are used in populations or joint or ensemble analyses.
In general, when multiple data point estimates are used jointly, the variance-induced offsets of the estimators average out but the bias-induced offsets remain fixed.
This is a straightforward point of statistics---this is not news---but it isn't currently informing most of the practice of ML in the natural sciences, where many catalogs and public data products are being produced with ML regressions \cite{spectrophotometric, leung, aspgap}, sometimes even by us.

In principle, there might be fixes for this problem that involve de-biasing the estimates.
This is not possible in general, because there is not usually enough validation data to accurately assess the bias.

In Bayesian language, this problem is closely related to the point that when data are to be combined, they should be combined at the likelihood level, not the posterior level.
Likelihoods are not affected by population-level biases.
Likelihoods can produce unbiased point estimates.
Likelihood-based estimators are not usually the lowest-risk estimators, but they can be low- or zero-bias estimators.
The outputs of regressions, on the other hand, are more like posterior-based estimators, affected by the implicit prior set by the training set.
When they are combined, the implicit prior gets amplified.
Most ML regressions are not capable of generating or emulating likelihoods.
There are interesting approaches in development to replace standard regressions with generative models that can produce likelihoods \cite{cannon, likelihood_ratio, sequential, biwei}.

Finally we note---and demonstrate in \appendixname~\ref{app:toy}---that none of these problems of bias arise from out-of-sample generalization or distribution-shift problems, or even model mis-specification.
Distribution shifts (where the test data are drawn from a different distribution than the training data) and model mis-specification (where the model doesn't have the expressivity to explain the data) just tend to make these biases even worse.

\paragraph{Emulator-induced confirmation bias:}
In many fields in the natural sciences, the fundamental theoretical model is \emph{computational}, meaning that theoretical predictions are made with large computer simulations.
These simulations are usually being asked to handle large ranges of length scales and time scales, so they are generally getting larger and more computationally demanding each year.
Quantitative comparison of data with simulations usually involves running enormous numbers of simulations \cite{abc}, because simulations have to span ranges of fundamental parameters and initial conditions and so on.
Thus these simulation requirements are getting very large for contemporary research projects.
For example, in cosmology, contemporary experiments could easily consume the total scientific computing capacity of the United States.
Hence we turn to emulators---trained ML regressions that have learned the input-output relationship of a simulation, or the relationship between an inexpensive low-resolution simulation and an expensive high-resolution experiment.
Emulators can make computationally impossible projects possible.

Recall that because these emulators are ML methods, they generally have uninterpretable internal parameters and weights, and they generally have been validated by comparison with held-out training data.
Like all high-capacity ML methods, they generally are vulnerable to adversarial attacks \cite{adversarial1}.

Now imagine that two cosmology experiments proceed to do enormous inferences on large data sets, each of which requires a very large number of simulations---so many that a fast, trained ML emulator has to be used.
Imagine that both of these inferences would have been exceedingly expensive to have been done with the original first-principles simulations instead of the emulators.
Experiment A finds cosmological parameters very much in line with our expectations going in.
Experiment B finds something extremely surprising about the mass matrix for the neutrinos, in conflict with other measurements and our expectations.
Is it possible that the anomaly discovered by Experiment B isn't real; it just comes from something very slightly wrong or off in the trained ML emulator?

By construction, both experiments are extremely expensive to re-analyze with first-principles simulations.
Which one would we fund for reproduction?
The motivation to re-analyze Experiment B is far, far higher than the motivation to re-analyze Experiment A.
That is a classic example of \emph{confirmation bias}.

Since the resources required are substantial (or even enormous), there is no simple fix for this unless we plan to re-analyze all experiments using first-principles simulations, no matter what they find.
If that is what's required to avoid the bias, then we should just analyze everything with the full simulations in the first place, and never use the emulators at any stage.

It is worthy of note that this problem could be solved, or greatly alleviated, if the inferences were performed with fully explainable methods, from beginning to end.
Right now there are no hopes for this, although as emulation methods are made to look more and more like physical models (see, eg, \citealt{scalars, villar2023dimensionless}) in their internal structure, maybe explainability will become possible.
This problem cannot be solved in general by making finitely many tests of the emulators.
First of all, such tests are expensive.
Second of all, in many cases (and in cosmology especially), the phase space of random initial conditions is so large that it is impossible to span or cover it in any finite set of tests.
Testing is good! But complete testing is impossible.

Finally, we note that one objection to this confirmation-bias point is that perhaps it exists just as strongly for any simulation-based inference (such as approximate bayesian computation; \citealt{abc})?
After all, simulations are expensive, so the full suite of them probably can't be affordably repeated.
There are two responses to this objection.
The first is that if the emulators work---that is, if they greatly reduce the cost of the inference---then the emulators have enormously increased the cost difference (or cost factor) between the inference and any re-analysis using physics-based simulations.
The second is that emulators are not doing precisely what the simulations did.
One way to see this is that they are subject to adversarial attacks \cite{melchior}.
Therefore we know that the emulator is not doing precisely what the physics-based simulation is doing.
That is, verification is not just more expensive, but also more important, when an emulator has been introduced.

\section{When is machine learning good for natural science?}\label{sec:good}
Where, in the natural sciences, can you use models in which you don't fully understand the latent parameters and structures?
The answer is: You can use them for the parts of your scientific project or experiment or analysis that \emph{you don't need to understand}.
We have colleagues who believe that there is no such part; some scientists believe that you must understand deeply every part of your scientific toolkit.
We disagree: For an example in physics, we use infrared HgCdTe detectors \cite{hgcdte} to make extremely precise measurements of intensity fields, without understanding in detail the solid-state physics that makes these devices efficient, linear detectors of photons.
There are many parts of a scientific project that must work well, but which do not need to be carefully controlled or understood.
They have to be testable and tested, but they don't have to be fully understood.

Here we give a few examples.
These are not intended to be exhaustive; they are just examples of places within a scientific project at which ML can be used to the benefit---and not detriment---of the natural-science goals.

\paragraph{Real-time execution and operations:}
In many scientific projects, experiments are adaptive to real-time outcomes, or data are taken selectively, or limited resources are assigned to particular targets.
When bandwidths are high, or experimental timescales are short, sometimes the decisions that need to be made in real time cannot be made with the first-principles decision-making that the investigator would like.
An example mentioned above is the selection of events for storage at the \textsl{LHC Atlas} Experiment \cite{atlas}.
An example from astronomy is the upcoming \textsl{Rubin Observatory LSST} project \cite{rubin}, which will generate some $10^5$ events of possible interest for follow-up observing every night \cite{lsst_events}.
These decisions can be made with very fast classifiers, trained on well-studied training data or simulations (eg, \citealt{lsst_broker}).

Is it unsafe to operate a project using ML decision-making systems?
Even if these systems do not have to be extremely well understood in real time, they do need to be \emph{versioned} and preserved for study later.
Statistical projects making use of the data from systems like this will need to understand the statistical biases created by the ML selection procedures.
This can be done after the fact, provided that the decision-making system is preserved in a precisely reproducible state, such that it can be run offline on counterfactual, simulated, and subsequent data.

\paragraph{Discoveries of outliers and rare objects:}
As mentioned in \secref{sec:why}, ML is well suited to outlier detection.
Since ML methods are expressive, and trained to fit the data, new data that \emph{aren't} well described by a trained ML model are possibly interesting.
Outlier-detection systems can be used to identify time intervals in which the equipment is malfunctioning.
They can also be used to identify objects that are rare or unusual.
Many discoveries in astrophysics are discoveries of new kinds of objects, originally found as data outliers \cite{quasars, voorwerp}.
So far, ML has not been involved in any big discoveries in astronomy, but we expect this to be a productive avenue for ML in the natural sciences (but see \citealt{contardo}).

\paragraph{Foreground, background, and confounder models:}
We mentioned in \secref{sec:philosophy} that a natural-science theory is a combination of a fundamental, latent theory with a (usually complicated) auxiliary observation model that explains the details of the observations.
The simplest part of this observation model is the model of the backgrounds, foregrounds, and confounders.
For example, any maps of the cosmic microwave background (by the ESA \textsl{Planck} spacecraft \cite{planck_maps}, for example) will have emission from the intervening Milky Way and from radio galaxies, which both add to and distort the maps.
When the goal is to understand the cosmic microwave background, the investigator often doesn't need to understand these foregrounds \emph{physically}; an \emph{effective model} is sufficient.
And indeed, all the bleeding-edge methods for this problem currently are ML methods \cite{cmb_foregrounds}.
It doesn't matter to the questions of physical cosmology what those foregrounds \emph{mean}; and indeed with ML-based foreground removal, cosmic microwave background experiments have delivered measurements of the parameters of the cosmological model with sub-percent precision \cite{planck_parameters}.
For foregrounds and backgrounds, understanding is not always necessary.

An even stronger statement can be made here, however, in the realm of causal inference:
Sometimes the goal is to argue that an effect is being caused by some particular cause, and there are confounding causal inputs that might be distorting or influencing the data.
The more expressive the model used to model the confounders, the stronger the conclusions that can be made about the causation \cite{causal_inference}.
That is, the huge capacity of ML methods can make a causal claim \emph{very conservative}, and hence very robust.
One way to see this is that if the confounder model has been given every chance it possibly can have to model the effect of interest, and it cannot, then it is a stronger conclusion that the effect is caused by the cause of interest.
This is the strategy in contemporary causal-inference projects, which make use of ML methods for these reasons \cite{bart}.

A key place for this kind of causal inference in astrophysics has been in instrument calibration.
Data taken by an instrument have signals imprinted from the physical (or biological or chemical) processes of interest, and also from the details of the measurement hardware, such as sensitivity and bias variations, and time-dependent distortions of internal mappings.
In the end, if the results of the investigation are to have the instrument calibrated out, it makes sense to give the instrument model a lot of expressive power, and judge that model in terms of its ability to explain the data.
Hence, instrument models and calibration are important places for ML methods, places where the ontology and epistemology of ML are well matched to the objectives, and places where the introduction of ML can made scientific projects more conservative and more accurate.
One very early example of this in astronomy was the calibration of the \textsl{Sloan Digital Sky Survey} imaging data \cite{ubercalibration}.

\section{Discussion}\label{sec:discussion}

We hope we have clearly answered the question in the title with the response ``Both!''
Our more detailed answers come from a perspective of physics, but we expect the answers to be similar in many of the natural sciences.
Machine learning (ML) is a reality for science:
We need it for speed and scale in many contexts in many contemporary scientific projects, where instrument bandwidths, data volumes, assay parallelization, experimental automation, and scientific ambitions are all growing.
It is also the case that PhD students and practitioners in the natural sciences want to---or even need to---learn ML practices to have useful and transferable skills.
ML is here, and it is here to stay.

Of course there is a certain amount of hype in this area, and we are seeing scientists ``move fast and break things'' all over the place.
If anything, this \documentname{} represents a call to slow down and think more.
ML tools are incredibly powerful, if we put them to use in the right places in the natural sciences.
In particular, the biases we highlight in \secref{sec:bad} do not have fixes; in our view ML simply cannot be used in those contexts without leading to substantial scientific mistakes.

There is a famous article by Wigner about ``the unreasonable effectiveness of mathematics in the natural sciences'' \cite{wigner}.
Mathematics has been pursued (in many cases) for abstract reasons of beauty and technical challenge, but it has ended up creating a language for many of the natural sciences.
Similarly, we could coin a phrase about \emph{the unreasonable effectiveness of machine learning in the natural sciences}.
It's unreasonable because current ML methods are typically engineered and optimized to solve problems in commercial applications.
The most remarkable ML systems have been driven by critical industrial questions such as:
Which advertisement should I display on this web search result?
What direction should I steer this automobile?
How could I make a chair that looks like an avocado?
And which online videos contain kittens?
In some sense it is remarkable that there is \emph{any} overlap between the technologies that solve these problems and the technologies we need in scientific domains.

But that said, the industrial successes of ML can also be misleading.
The fact that big ML models can solve countless problems in business does not mean that all areas of the natural sciences will be helped by the introduction of ML.
You do not have to \emph{understand} your customers (nor their videos) in order to make plenty of revenue off of them; in the natural sciences, it is understanding, not revenue, that we seek.

One scientific-practice comment to make is that the ML literature is very different from the natural-science literatures.
In ML, the publishability of a result is decided---or very strongly influenced---by its demonstrated performance on held-out data.
It is almost impossible in ML to publish contributions that do not lead to the immediate improvement of some prediction.
This has a positive side: There are standards!
But it has many negative sides.
For example, the path from one good method to another good method might necessarily pass through a valley of less-good methods.
These non-local-optimization paths aren't available to the ML literature at present.
Whether good or bad, it is interesting to us that publishing in ML is very different from publishing in the sciences.

Given everything we have said here, what are the ways forward for the natural sciences?
ML is critical in many projects; how do we make sure it is doing what we want it to do?
Our view is that the promising directions include imposing physically motivated mathematical structure into the design or regularization of the ML models.
This includes symmetries, conservation laws, and constraints \cite{bronstein2021geometric, villar2023towards}, but also ODE and PDE solvers \cite{karniadakis2021physics}, dynamic programming modules \cite{xu2020can}, and other established mathematical tools.
The research direction that looks towards explainability, interpretability \cite{interpretable}, and trustworthiness \cite{kearns2019ethical} is also fundamental to the development of ML methods for the physical sciences.
These approaches can increase natural scientists’ confidence in ML tools, and also perhaps expand their ontologies and epistemologies.

Finally, we make a philosophical comment:
In the natural sciences, the same phenomena can usually be explained with many qualitatively different theories; theories are not usually unique (eg, \citealt{hogg, peebles}).
In the same way (as we noted above), ML models contain massive degeneracies, and even different architectures can lead to near-identical predictions after training on the same training data.
Does this lead to a philosophical connection between ML and the natural sciences?
We have argued previously \cite{hogg} that the degeneracies in the natural sciences undermine \emph{realism} or the belief in the literal truth of the latent theories or objects of physics.
There is an argument to be made that maybe the natural sciences could be better off if they moved at least slightly closer to the ML points of view on ontology and epistemology.
In the long run, the data are more stable than the fundamental theories of physics:
Newtonian gravity was replaced with general relativity, and we expect general relativity to be replaced by some kind of quantum gravitational theory.
At each change, the ontology changes, while the data are stable.
We are not recommending anything (yet), we are merely noting that a robust philosophy of natural science might be more positively oriented towards the data than current standard practice.

\paragraph{Acknowledgements:}
It is a pleasure to thank
  Gaby Contardo (SISSA),
  Jennifer Hill (NYU),
  Adrian Price-Whelan (Flatiron),
  Hans-Walter Rix (MPIA),
  Sam Roweis (deceased), and
  Bernhard Sch\"olkopf (MPI-IS)
for conversations over the years related to these arguments, and
  Vedant Chandra (Harvard),
  Contardo,
  Valentino Foit (NYU),
  Alex Jiang (CUNY),
  Charles Margossian (Flatiron),
  Rix,
  Ilya Shpitser (JHU),
  Kate Storey-Fisher (DIPC),
  Jeremias Sulam (JHU),
  Jesse Thaler (MIT),
  Sebastian Wagner-Carena (Flatiron),
  Kaze Wong (Flatiron), and
  our four anonymous ICML reviewers
for valuable comments on this particular manuscript.
SV was partially supported by ONR N00014-22-1-2126, NSF CCF 2212457, the NSF–Simons Research Collaboration
on the Mathematical and Scientific Foundations of Deep Learning (MoDL) (NSF DMS 2031985), and NSF
CAREER 2339682.
The Flatiron Institute is a division of the Simons Foundation.

\section*{Impact statement}
This \documentname{} is directly aimed at producing certain kinds of broader impacts, in the execution of natural-science investigations and projects.
ML has many roles in the natural sciences.
This work is intended to start a discussion which, in time, will support the implementation of ML in good roles in scientific projects, and criticize the implementation of ML in less good roles in scientific projects.
It is also intended to help practitioners understand and respond to biases introduced by the use of ML in those less-good roles.
The goal of this work is to improve the accuracy and success of the natural sciences.
However, if the recommendations in this \documentname{} are adopted by scientific communities, the costs of some scientific projects may increase.

\bibliography{main}
\bibliographystyle{icml2024}

\clearpage\appendix
\section{Population-level biases in regression outputs}\label{app:toy}
A standard machine-learning (ML) regression learns a function $f(x;\theta)$ that takes as input a feature list or vector $x$ and outputs a label estimate $y$.
In detail, the method learns the value of a list or vector $\hat{\theta}$ of parameters (weights and thresholds, for example) of the function that leads to the best predictions of the labels in some training set; the training set is a set of labeled data, or a set of $(x, y)$ pairs.
The learned function $f(x;\hat{\theta})$ delivers a predicted or estimated label $\hat{y}$ for any data point with a complete set of features $x$.
Thus a particular estimated label $\hat{y}_\ast$ for a particular data point with features $x_\ast$ obtains information \emph{both} from the features $x_\ast$ and \emph{also} from the features and labels of all the data points in the training set used to set the parameters $\hat{\theta}$.
In applications in which regression-estimated labels are used, it matters \emph{how much information} is coming from the individual data-point's features and how much from the original training set.

\begin{figure*}[p!]
\includegraphics[width=0.49\textwidth]{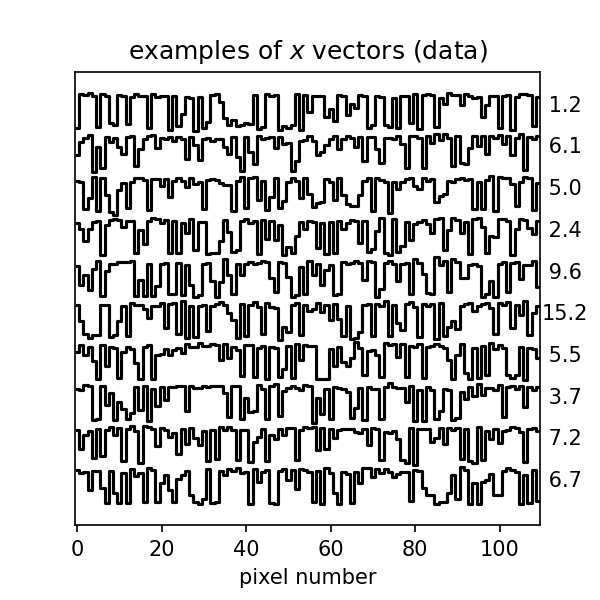}
\includegraphics[width=0.49\textwidth]{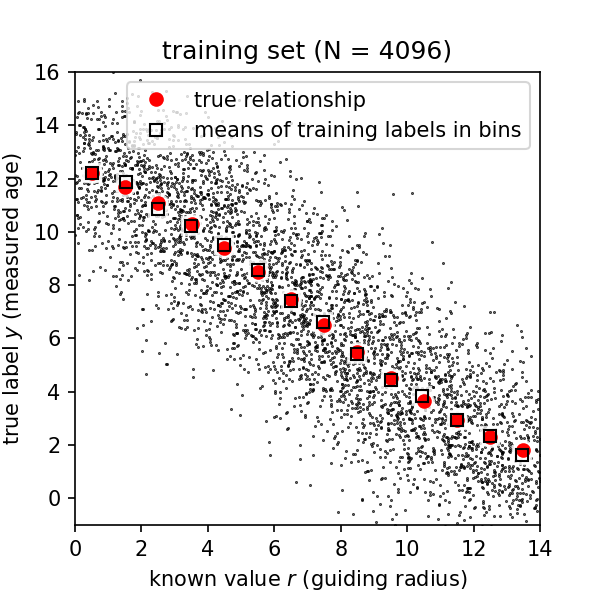} \\
\includegraphics[width=0.49\textwidth]{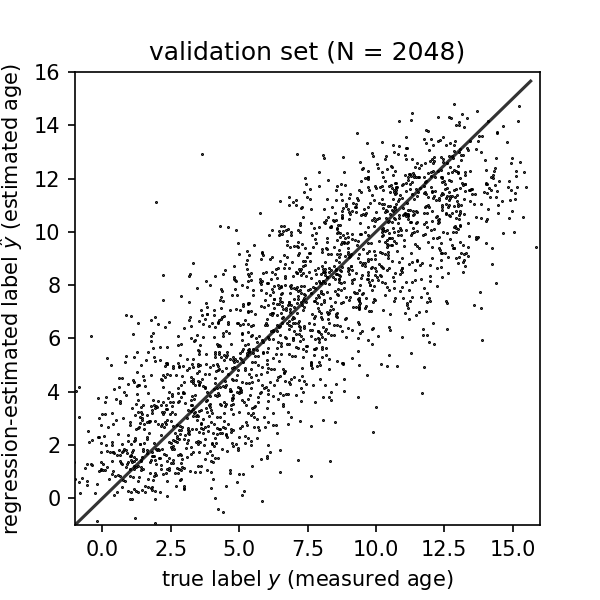}
\includegraphics[width=0.49\textwidth]{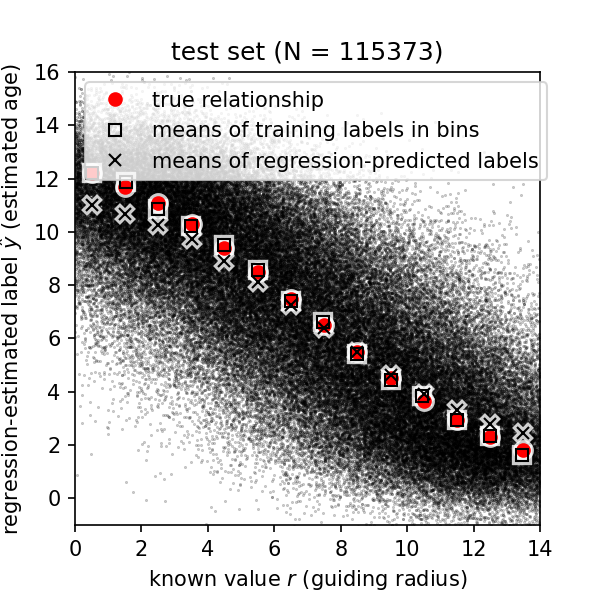}
\caption{
Visualization of the toy regression. \textsl{Top-left:} Random examples of the data vectors $x$, which are one-dimensional images generated from a linear model plus a nonlinearity created by two rectifications. Details of the data generation are given in the text. Each example $x$ is labeled on the right side by the value of its label $y$. \textsl{Top-right:} The training-set labels $y$, plotted against the known parameter $r$, which is not used in the regression (only the vectors $x$ are used). Also shown are solid red circles showing the true mean relationship between $y$ and $r$ in the toy data. Open black squares show the empirical mean relationship measured in bins in the training-set data. \textsl{Bottom-left:} Validation of the trained regression in the validation set, showing that the label estimates $\hat{y}$ are noisy (as expected given the problem set-up) but not strongly biased. \textsl{Bottom-right:} The regression estimates $\hat{y}$ in the very large test set, plotted against the known parameter $r$. Also shown are the same solid red circles and open black squares as in the top-right plot. Black X-shaped symbols show the mean relationship between $\hat{y}$ and $r$. The relationship shown by the Xs is very precise (error bars are much smaller than the symbols; see the text) but biased far away from the true relationship, unlike the relationship shown by the open squares (measured in the training set alone).\label{fig:regression}}
\end{figure*}

To illustrate these ideas, we construct a toy data analysis.
This toy will make a general point, but it has a decidedly astronomical feel to it.
The complete toy data set is generated by a process with the algorithm below.
This algorithm makes references to objects $\xi_\eta$ and $\xi_\zeta$ which are generated prior to the start.
\begin{hoggnumerate}
    \item A floating-point value of a known parameter called ``guiding radius'' $r$ is generated in the range $0<r<14$.
    \item A floating-point value of a latent parameter called ``age'' $\eta$ is generated from a Gaussian with mean $14 - r$ and variance 4.
    \item The point is discarded if the age is outside the range $0<\eta<14$. If the point is not discarded:
    \item A $K$-dimensional latent vector $\zeta$ with $K=14$ is generated from a Gaussian with zero mean and variance 1.
    \item A $M$-dimensional latent vector $\xi$ with $M=110$ is created by $\xi = \xi_\eta\,\eta + \xi_\zeta\cdot\zeta$, where $\xi_\eta$ is a $M$-vector and $\xi_\zeta$ is a $M\times K$ matrix.
    The elements of $\xi_\eta$ and $\xi_\zeta$ were drawn (before the start) from Gaussians with zero mean and variances $0.01/K$ and $1/K$ respectively.
    \item A rectified latent vector $\tilde{\xi}$ is created by taking $1 + \xi$ and rectifying all pixels to lie in the range $[0, 1]$.
    \item A $M$-dimensional feature vector $x$ called ``the data'' is generated from a Gaussian with mean $\tilde{\xi}$ and variance $0.0025$. (These feature vectors have been designed to look a little like stellar spectra.)
    \item A label $y$ called ``measured age'' is generated from a Gaussian with mean $\eta$ and variance 1.
\end{hoggnumerate}
The data generated this way are shown in the top panels of \figref{fig:regression}.
The data contain a non-trivial relationship between label $y$ (measured age) and known parameter $r$ (guiding radius).
This relationship is shown in \figref{fig:regression}.
It isn't precisely linear because of the removal of data points with latent ages $\eta$ outside the range $0<\eta<14$.

We construct a three-layer multilayer perceptron (MLP) (using the scikit-learn implementation) with layer sizes of 64, 32, and 16 neurons.
The model is trained on a training set of 4096 data points with features $x$ and labels $y$.
The trained model takes as input features $x$ and outputs estimated labels $\hat{y}$.
The model is validated on a validation set of 2048 data points with features $x$ and labels $y$.
The test set is a much larger set of $10^5$ data points with features $x$ and no labels.
The model is used to make estimated labels $\hat{y}$ in the full test set.
The validation and test labels are shown in the bottom panels of \figref{fig:regression}.
Importantly, \emph{the training, validation, and test samples are drawn from exactly the same process}, with the same distribution in all properties.

The regression-estimated labels $\hat{y}$ are related to the measured labels $y$ in the held-out validation set with a linear relationship with unit slope and zero intercept.
The comparison to the validation data (bottom-left panel of \figref{fig:regression}) shows that they are not precisely estimated, but there is no evidence for strong bias at the individual-object level.

The true relationship between label $y$ (measured age) and known parameter $r$ (guiding radius) is estimated in a large generated data set by taking means of $y$ in bins of $r$.
This relationship is shown in \figref{fig:regression} with solid red circles.
The empirical relationship between label $y$ and known parameter $r$ in the training-set data is shown in \figref{fig:regression} with open black squares.
These two relationships are statistically consistent, as expected.
The empirical relationship in the test-set data between the estimated label $\hat{y}$ and the known parameter $r$ in the test-set data is shown in \figref{fig:regression} with black X-shaped symbols.
The test-set relationship is strongly biased, deviating from the true relationship at the edges of the $r$ range.

In this example, not only is the test set providing a biased answer (to the question of interest), while the training set is not, it is also the case that the test-set answer is delivered with \emph{very high confidence}.
Because of the number of objects involved in the test set, the mean-age values are computed at very high precision (the square root of $N$ is large).
Thus it is not just the case that the test-set answer is biased, it is biased at the 30-sigma level away from the truth.
The training set gives less precise answers (it is smaller), but the answers are consistent with the truth.

One objection to this example is that the information content in the features about the label is low.
Indeed! This situation is common in stellar spectroscopy, where we care about subtle abundance and age effects.
The bias does indeed go down as the information content in the features goes up.

Another objection is that maybe this all depends on model capacity or expressivity.
The data in this case are made so simply that the regression being used contains sufficient expressivity to model the data, so that issue isn't dominating this problem.
However, in detail, the variance and bias of the ML regression estimator does depend on the effective model space, and sometimes in problems like this, simpler (less expressive) models can be less biased (see, eg, the Gauss-Markov theorem).
Full tests of this are out of scope here.

In this case, what should the investigator do---the investigator who wants to know the relationship between $y$ (age) and $r$ (guiding radius)?
The investigator should \emph{just use the training set}.
Transferring labels to the test set did not help---indeed it actually hurt---even though the test set is far larger than the training set, and drawn from the same distribution.
The test-set results are very precise (error bars on the X-shaped symbols cannot be visibly shown in \figref{fig:regression}), but they are very wrong.
The training-set results are less precise but unbiased; they are much closer to the truth.

Indeed, none of the issues here come from out-of-distribution problems or distribution shifts.
All three data sets (train, validate, and test) are drawn from an identical parent population.
When distribution shifts are involved (usually the labeled data are the best data, or the earliest, or the easiest to obtain), these kinds of problems only get worse.

Related to that: The most interesting aspect of this problem is that there is no \emph{wrong} bias:
The bias introduced from the training set is a correct bias; the training set represents a prior that accords with our beliefs about the test set.
And yet the population-level inference is biased.
The averaging of points shown by the X-shaped symbols in \figref{fig:regression} is a simple misuse of the regression outputs:
The averaging of biased estimates amplifies the statistical significance of the estimator's biases.
However, it is a misuse that is occurring repeatedly in astrophysics with catalogs generated by ML regressions.

Another way to say it is:
If you expect your downstream users to perform populations inferences with your outputs, you want your outputs to be unbiased.
The methods you choose must be informed by the subjective needs and expectations of your users.
ML regressions aren't appropriate to the use cases of many downstream users.

All code (including the data-generating code) used in this toy are available under an open-source license at 
\url{https://github.com/davidwhogg/BadForScience}.

\end{document}